\RequirePackage{fix-cm}
\documentclass[letterpaper]{article}
\usepackage{aaai}
\usepackage{times}
\usepackage{helvet}
\usepackage{courier}
\usepackage[T1]{fontenc}    
\usepackage{hyperref}       
\usepackage{url}            
\usepackage{booktabs}  
\usepackage{graphicx}
\usepackage{amsfonts}       
\usepackage{nicefrac}   
\usepackage{subfigure}
\usepackage{microtype} 
\usepackage{amsmath}
\usepackage{algorithm}
\usepackage{algpseudocode}
\usepackage{wrapfig}
\usepackage{bm}
\usepackage{enumitem}
\usepackage{multirow}
\usepackage{xcolor}

\newcommand{\sn}[1]{\textcolor{purple}{[#1 \textsc{--Sriraam}]}}
\newcommand{\D}{\mathcal{D}}
\newcommand{\M}{\mathcal{M}}
\newcommand{\F}{\mathcal{F}}
\newcommand{\T}{\mathcal{T}}

\newcommand{\thetab}{{\bm \theta}} 

\newcommand{\pred}[1]{\ensuremath{\mathtt{#1}}}

\floatname{algorithm}{Procedure}
\renewcommand{\algorithmicrequire}{\textbf{Input:}}
\renewcommand{\algorithmicensure}{\textbf{Output:}}
\begin{document}
%
\title{Non-Parametric Learning of Gaifman Models}
\author{Devendra Singh Dhami, Siwen Yan, Gautam Kunapuli and Sriraam Natarajan\\
The University of Texas at Dallas\\
\{devendra.dhami,siwen.yan,gautam.kunapuli,sriraam.natarajan\}@utdallas.edu}

\maketitle
\begin{abstract}
\begin{quote}
We consider the problem of structure learning for Gaifman models and learn relational features that can be used to derive feature representations from a knowledge base. These relational features are first-order rules that are then partially grounded and counted over local neighborhoods of a Gaifman model to obtain the feature representations. We propose a method for learning these relational features for a Gaifman model by using relational tree distances. Our empirical evaluation on real data sets demonstrates the superiority of our approach over classical rule-learning.
\end{quote}
\end{abstract}

\section{Introduction}

Learning embeddings of large knowledge bases has become a necessity due to the importance of reasoning about objects, their attributes and relations in large graphs. Statistical Relational AI/Learning (StaRAI)~\cite{raedt2016statistical,getoor2007introduction}, has the ability to learn and reason with multi-relational data in the presence of uncertainty. 
While specific models such as Markov Logic~\cite{richardson2006markov}, ProbLog~\cite{de2007problog} and PSL~\cite{brocheler2010probabilistic} (to name a few) exist, a more scalable model~\cite{niepert2016discriminative} was proposed recently. 
This work built on Gaifman's locality theorem~\cite{gaifman1982local,grohe2004existential}, which states that every first-order sentence is equivalent to a Boolean combination of
sentences whose quantifiers range over local neighborhoods of the Gaifman graph. 
The key idea is that if one could identify effective representations from local neighborhoods (of objects or tuples of objects), one could learn machine learning models that can be used for reasoning in large graphs. This ``local representation'' approach was inspired by the success and scalability of convolutional neural networks (CNNs, \cite{Goodfellow-et-al-2016}), specifically, the ability of CNNs to engineer complex features from locally-connected image neighborhoods. 

In a similar manner, relational Gaifman models seek to identify locally-connected relational neighborhoods within knowledge bases for effective representation, learning and inference. 
While effective, the relational learning model recently proposed by Niepert \cite{niepert2016discriminative}, called {\bf Discriminative Gaifman Models}, used relational features that were hand-crafted rather than learned, that is, structure learning (to use the terminology from probabilistic graphical models) was not performed.

We address this problem of structure learning \footnote{This is a modified version of the paper presented at PLP 2019. We are submitting to StarAI to solicit feedback from the community.}: learning relational features for training the Gaifman model. We consider three different approaches. (1) As suggested by Niepert~\cite{niepert2016discriminative}, we employ Inductive Logic Programming (ILP)~\cite{muggleton1991inductive} to learn discriminative first-order rules; 
(2) Inspired by the success of random walks in deep relational models ~\cite{lao2010relational,lao2011random,kaur2017relational}, we employ relational random walks; 
(3) Finally, as a novel contribution, we propose the use of paths from relational trees learned via relational one-class classification~\cite{khot2014relational};  specifically, each path from root to leaf of a relational tree can be considered a relational feature. Given these relational features, we apply traditional discriminative machine learning algorithms. 

 We make the following key contributions. (1) We present a method for {\em learning relational embeddings} for reasoning over large graphs. (2) We adapt a recently developed relational learning method for constructing relational features. (3) We adapt well-known relational rule learners for learning local neighborhood representations. (4) We combine these relational features with discriminative classifiers to learn discriminative Gaifman models. (5) We demonstrate that combining the more novel relational trees with a discriminative classifier is more effective in learning on large graphs compared to a standard ILP learner. (6) Our empirical evaluation reveals an important characteristic of our approach:  high recall without sacrificing precision in both medical and imbalanced data sets. {\bf This is the first work on structure learning for Gaifman models}. 
\begin{figure*}[!tb]
    \begin{center}
    \begin{minipage}[t]{0.35\linewidth}
    \vspace{0pt}
    \includegraphics[width=\columnwidth]{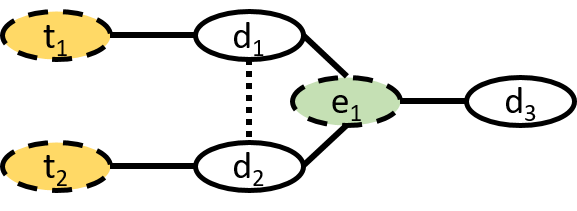}
    \end{minipage}
    \begin{minipage}[t]{0.6\linewidth}
    \vspace{0pt}
    {\small
    \begin{tabular}{l}
         \hline
        $\mathtt{TransportSubstr(Pravastatin, \, BileSaltExportPump)} $ \\ $\mathtt{TransportInhib(Simvastatin, \, MultidrugResistProtein1)} $\\
        $\mathtt{EnzymeInhib(Pravastatin, \, CytochromeP4502C9)} $ \\ 
        $\mathtt{EnzymeSubstr(Acetaminophen, \, CytochromeP4502C9)} $\\
        $\mathtt{EnzymeInhib(Simvastatin, \, CytochromeP4502C9)} $\\
        \hline
    \end{tabular}}
    \end{minipage}
    \hspace{0.1in}
    \end{center}
    \vspace{-0.1in}
    \caption{\small An example Gaifman graph for a drug-drug interaction (DDI) knowledge base. Here $d_1,d_2,d_3$ = \{Pravastatin, Simvastatin, Acetaminophen\}, $t_1$=\{Bile salt export pump\}, $t_2$=\{Multidrug resistance protein 1\} and $e_1$=\{Cytochrome P450 2C9\}. Note that the dotted line between $d_1$ and $d_2$ is the link we want to predict.}
    \label{fig:gaif}
    \vspace{-0.25in}
\end{figure*}

\section{Background and Related Work}
A \textbf{grounding} of a predicate with logical variables $x_1, \dots, x_k$ is a substitution $\{\langle x_1, \dots, x_k \rangle \slash \langle X_1, \dots, X_k \rangle\}$ mapping each of its variables to a constant in the population of that variable. 
A {\bf knowledge base} $\mathcal{B}$ consists of (1) a finite domain of objects $\D$ (also known as entities), (2) a set of predicates $\mathcal{R}$ that describe the attributes and relationships of the objects, and (3) an interpretation that assigns a truth value to every grounded predicate.

\paragraph{Discriminative Gaifman Models:}
The Gaifman graph $\mathcal{G}$, also known as the primal graph, of a knowledge base $\mathcal{B}$ is an {\bf undirected graph}, where the nodes are the entities $e \in \D$. $\mathcal{G}$ contains edges joining two nodes only if the entities $a$ and $b$ corresponding to those nodes are present in a relation together $\pred{R}(\hdots, \, a, \, \hdots, b, \, \hdots) \, \in \, \mathcal{B}$. $\mathcal{G}$ can be used to easily identify co-occurrences (or lack thereof) among every pair of entities in $\mathcal{B}$. Furthermore, cliques in $\mathcal{G}$ group entities that co-occur pairwise through shared relationships, and such cliques capture the local structure of a knowledge base. We illustrate this in Figure \ref{fig:gaif}, which shows a knowledge-base fragment and the corresponding Gaifman graph for drug-drug interaction (DDI). Given entities (drugs, enzymes, transporters) and relations between them, the underlying machine-learning task is to predict if two drugs interact. The dotted line represents the target predicate, and identifying it is {\bf link prediction}.

\begin{wrapfigure}{l}{0.5\columnwidth}
    \centering
    \includegraphics[width=0.48\columnwidth]{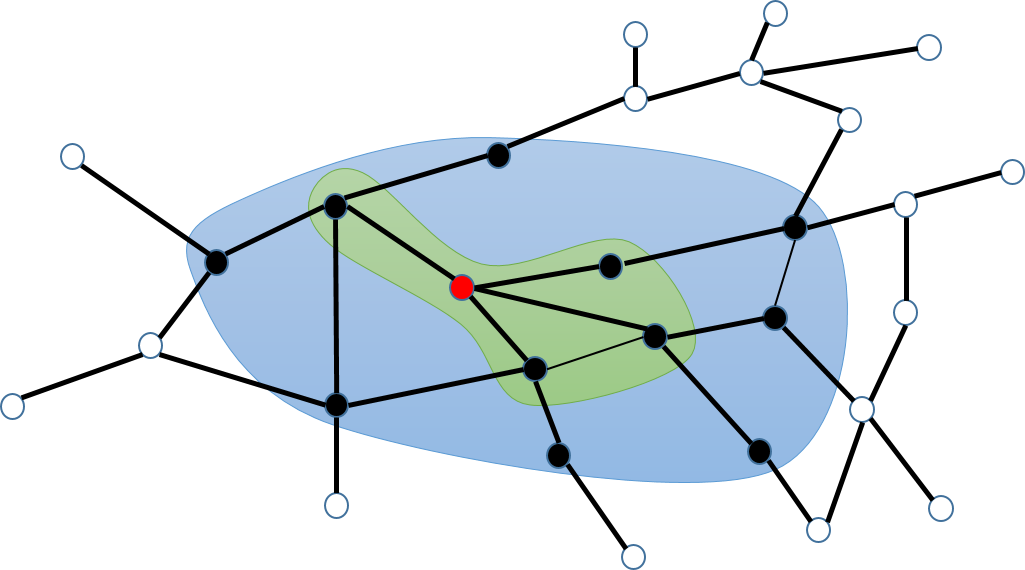}
    \caption{Gaifman neighborhoods.}
    \label{fig:neigh}
\end{wrapfigure}

The distance $d(a, \, b)$ between two nodes $(a, \, b) \in \mathcal{G}$ is the minimum number of hops required to reach node $b$ from node $a$. For example, in Fig. \ref{fig:gaif},  $d(d_1, \, t_1) \, = \, 1$ and $d(d_1, \, d_2) =2$. The $r$-neighborhood of a $a \in \mathcal{G}$ is the set of all nodes that are at most a distance $r$ from $a$ in the Gaifman graph: $N_r^\mathcal{G}(a)= \{\bar{a} \in \mathcal{G}\, \mid \, d(a,\bar{a}) \leq r \}$. For example, $N_1(d_1) = \{t_1,e_1\}$ and $N_2(d_1) = \{t_1,e_1,d_2,d_3\}$. Figure \ref{fig:neigh} shows the 1 and 2-neighborhood of a node (colored red) in a given Gaifman model. When a first-order rule/clause $\varphi(x)$ is relativized by the neighborhood of the free variable $x$, the resulting first-order rule $\psi^{N_r(x)}(x)$ is called $r$-local. A Gaifman neighborhood can be thought of as representing second-order proximity between nodes. The interpretation is that nodes with shared neighbors are more likely to be similar and more likely to have a link between them. 
Discriminative Gaifman Models (DGMs, \cite{niepert2016discriminative}) are relational models that can exploit structural features of a local neighborhood of a knowledge base. These structural features are aggregated from locally-sampled neighborhoods, and the aggregation is based on the \textit{Gaifman locality theorem} \cite{gaifman1982local} stated as: 
{\em Every first-order sentence is logically equivalent to a Boolean combination of basic $r$-local sentences}.
An $r$-local sentence is of the form $\exists x_1 \, \hdots \, \exists x_k \,\,  \left( \bigwedge_{1 \leq i < j \leq k} \textit{d}(x_i,x_j) > 2r \,\, \land \,\, \bigwedge_{1 \leq i \leq j} \varphi(x_i) \right)$, where $r, \, j \geq 1$ and $\varphi$ is an $r$-local first order formula.
In simpler terms, the locality theorem states that only a small part of a given structure is relevant for evaluating a query statement and thus a global structure search is not required. For example, if querying about the drug $d_1$ in Figure \ref{fig:gaif}, a search within the $1$-neighborhood of $e_1$ (say), that is $\{t_1, \, e_1\}$ is more relevant than searching through the complete graph which can be greatly computationally inefficient. Another way to look at the theorem is: \textit{a first-order rule is true if it is true in the local $r$-neighborhoods of a given graph}.
The DGM approach uses the Gaifman locality theorem to generate low-level embeddings for a given knowledge graph, which can then be used as propositional features in a standard classifier. Learning embeddings from a given knowledge graph, or {\bf graph embedding},  is a well-studied problem in machine learning. A large body of recent work in this area can be categorized broadly by the underlying approaches: matrix factorization, deep learning, edge reconstruction, graph kernels and generative models. These approaches have been extensively surveyed recently; see for instance, Nickel et al. \cite{NickelEtAl15}, Wang et al. \cite{WangEtAl17} and Cai et al. \cite{CaiEtAl18}. In general, Gaifman models tend to scale better than many such approaches to higher-arity relations and target-query complexity \cite{niepert2016discriminative} owing to their local view and incorporation of count-based features. 


\textbf{Relational and Structure Learning:}  One of the most important tasks in relational learning is that of \textit{link prediction} which determines whether a relation (link) exists between entities based on the given relational database. Taskar et al. \cite{taskar2004link} use template graphical models. Martinez et al. \cite{martinez2017survey} and Hasan et al. \cite{al2011survey} present a comprehensive survey on link prediction problems in complex networks and social networks respectively. Graph neural networks \cite{harada2018dual}, metric-learning~\cite{chuan2018link} etc. have been used for link prediction.
Structure learning in relational probabilistic models can be interpreted as learning relational rules from the data. In the context of ground Bayes nets, several methods such as genetic algorithms \cite{larranaga1996structure}, linear programming \cite{jaakkola2010learning} and constraints \cite{de2009structure} have been explored. For bi/undirected graphical models, local learning can be facilitated using gradient-boosting~\cite{khot2011learning}. 

\section{Learning Discriminative Gaifman models}

\begin{figure*}[!t]
    \centering
    \includegraphics[width=0.9\textwidth]{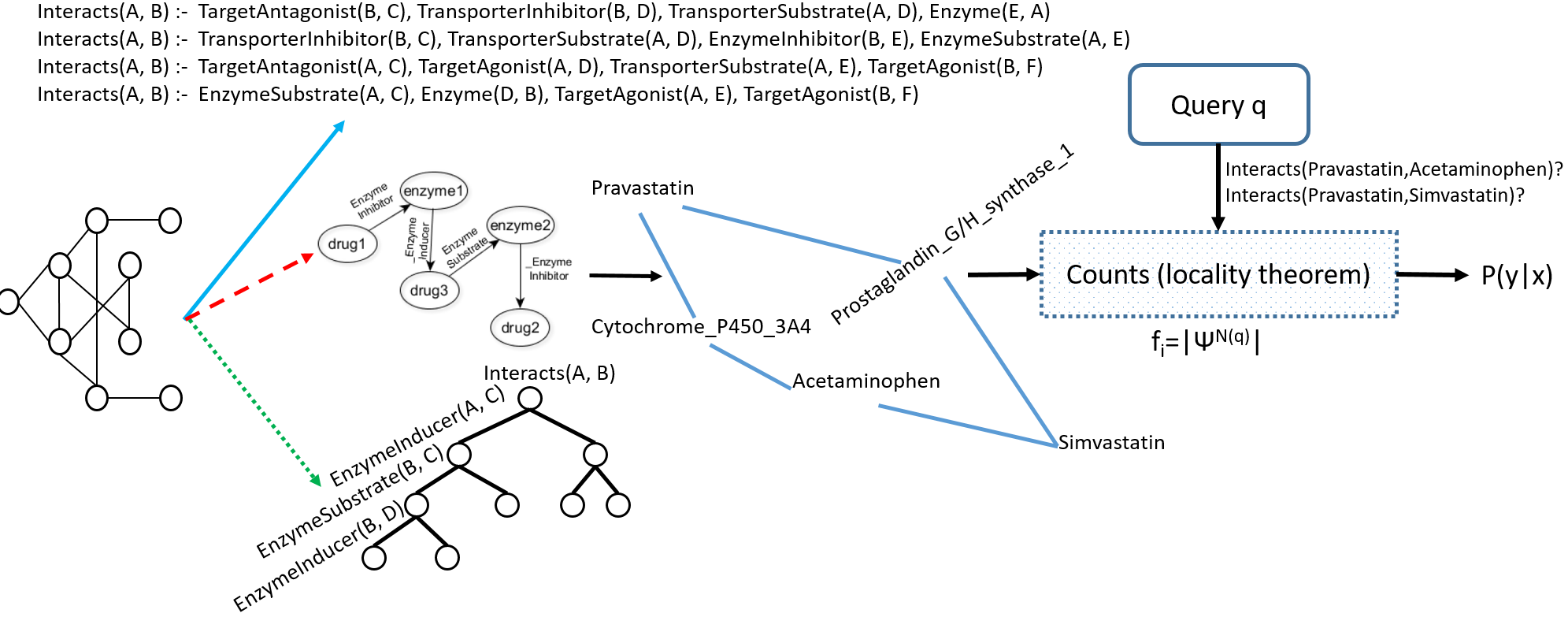}
    \caption{\small A general overview of link prediction using Gaifman models.}
    \label{fig:overview}
\end{figure*}

 \noindent \fbox{
 \parbox{\columnwidth}{
 \textbf{Given:} A knowledge base $\mathcal{B}$, facts $F_s$, and its corresponding Gaifman graph $\mathcal{G}$; \\
 \textbf{Output:} A DGM $\M$ that is trained for a particular link prediction task $\T$; \\
 \textbf{To Do:} Construct a set of relational features $\Phi$, and train a discriminative learner to predict $\T$.
 }}\\
Our approach, {\em Learning Gaifman Embeddings} (LGE), (1) constructs rules $\Phi$ that form the base set of relational features; (2) instantiates rules and performs counting based on task $\T$ to construct propositional features $\F$; and finally, (3) learns a discriminative classifier with $\F$ (Figure \ref{fig:overview}). 

\vspace{-0.5em}
\subsection{Learning Relational Rules}
\label{subsec: relational features}
\vspace{-0.5em}
Given a knowledge base $\mathcal{B}$, the Gaifman graph $\mathcal{G}$ is obtained by instantiating the entities that are connected by an edge type (relation) together in the form $\mathtt{R(e_1, \, e_2)}$, that is, $relation(type_1, \, type_2)$. The relation (link) to be predicted, defined by the target predicate, forms the set of positive examples. We make the {\em closed-world assumption}, that is, unobserved edges in the graphs are negative examples. 
Each relational example also has {\em facts} associated with it, which are the ground predicates in $\mathcal{B}$ that describe relational example, its attributes and relationships. All such facts are denoted $F_s$.


{\bf Features via Relational Rule Learning}. Our first solution is inspired by Niepert~\cite{niepert2016discriminative}, who suggested the use of an Inductive Logic Programming (ILP) style learning method. This method learns a set of discriminative Horn clauses (implications of the form {\em if <condition> then <consequence>}). Specifically, we use an ILP system called WILL ~\cite{natarajanilp} to learn the relational features\footnote{Any other ILP learner such as \cite{srinivasan2001aleph}, \cite{quinlan1990learning} or \cite{muggleton1995inverse} could be used.}. WILL first selects an example from the set of all examples and then finds a clause (rule) that \textit{best covers} the examples. The \textit{best covering }is the most general clause that maximizes the difference between the number of positive and negative examples covered
\footnote{Ideal coverage means {\em all} positive examples and {\em no} negative examples which can easily overfit.}. 
Each \textit{best covering} clause becomes a relational rule in our model. The examples covered by the clause are then removed and the process is repeated till a stopping criterion is satisfied;. 
for example, we have extracted a maximum number of rules/clauses. Note that when a stopping criterion requires $n$ rules to be extracted, it is sometimes possible to extract $m < n$ rules that cover the examples adequately. 

{\bf Features via Relational Random Walks}. Relational data is often represented using a graph that defines a domain's schema; in such a representation, a relation $\mathtt{R(e_1, e_2)}$ is a predicate edge between two entity type nodes: $\mathtt{e_1 \, \xrightarrow{R} e_2}$. A relational random walk (RW) through a graph is a chain of such edges corresponding to a conjunction of predicates. For a random walk to be semantically sound, we should ensure that the input type (domain) of the $i+1$-th predicate is the same as the output type (range) of the $i$-th predicate. An example relational random walk from the drug-discovery domain is:
\[
\begin{array}{cl}
& \mathtt{Interacts(d0, d3)} \, \Leftarrow \, 
\mathtt{TargetInhib(d0, t0)}\\
& \wedge \mathtt{\_TargetInhib(t0, d1)}
 \wedge \mathtt{TransporterSubstr(d1, t2)}\\
& \wedge \mathtt{\_TransporterInhib(t2, d3)}. 
\end{array}
\]
This is a semantically sound random walk as it is possible to chain the second argument of each predicate to the first argument of the succeeding predicate. This random walk also contains {\em inverse predicates} (prefixed by an underscore, such as $\mathtt{\_Transporter}$). Inverse predicates are distinct from their corresponding predicates as their arguments are {\em reversed}. Thus, this relational random walk chains the first variable $\mathtt{d0}$ in the target predicate $\mathtt{Interacts(d0, d3)}$  with the second variable $\mathtt{d3}$. The chain represents a relational feature and constitutes a random local structure of the form:
\begin{eqnarray*}
\mathtt{d0 \xrightarrow{TargetInhibitor} t0 \xrightarrow{\_TargetInhibitor} }\cdots \\
\mathtt{d1 \xrightarrow{TransporterSubstrate} t2 \xrightarrow{\_TransporterInhib}d3}. 
\end{eqnarray*}
Thus, to construct a relational random walk, only the schema describing the knowledge base is required. We adapt path-constrained random walks (PCRW, \cite{lao2010relational}) to construct relational random walks. The algorithm starts at the first entity in the target relation, and makes a walk over the (parameterized) graph to end at the second entity present in the target relation. One limitation of PCRW is that the random walks are only performed over binary relations. However, since we employ a predicate representation, {\em we generalize 
and learn with arbitrary $n$-ary relations}. 




{\bf Features via Relational One-Class Classification (relOCC) Features}. A common issue in many real-world relational domains, especially knowledge bases, is that only ``positive'' instances of a relation are annotated, while ``negative'' instances are not explicitly identified. This is because the number of instances where the relation does not hold is very large, and annotation can be prohibitively expensive. Learning with highly imbalanced data sets requires reasoning over just the positive instances, commonly referred to as {\em one-class classification}. Intuitively, if we can construct a relational one-class classifier describing the positive examples, then rules characterizing this classifier are essentially features that describe positive examples.
One-class classification typically requires a {\em distance measure} to characterize the density of the positive class. While, for standard vector and matrix data, many different distance measures exist, the issue is far more challenging for relational data, and depends on the underlying representation of the classifier.
\begin{figure}[t]
\caption{\small An illustration of Least common ancestor.}
        \centering
        \hspace{-0.1in}
        \includegraphics[width=0.8\columnwidth]{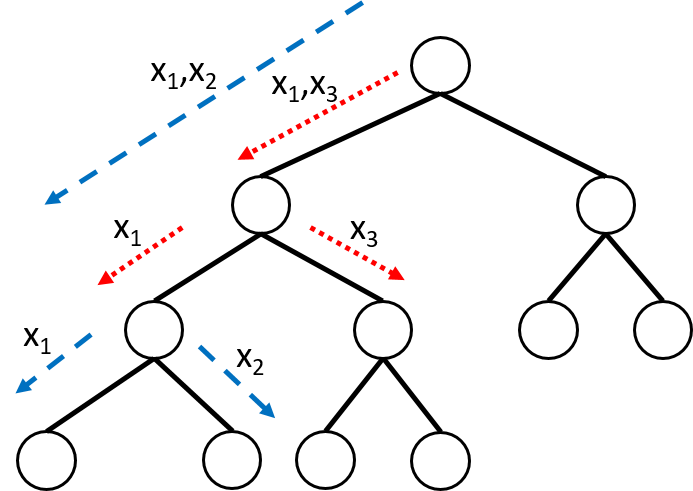}
        \label{fig:lca}
\end{figure}

Suppose we use an off-the-shelf learner to learn relational trees \cite{BlockeelDeRaedt98} to describe each class in the data. Such relational trees form a decision-list of relational rules. These trees can then be used to compute the {\em relational distance} between a pair of examples $x_1$ and $x_2$, 
\begin{equation}
    d(x_1, x_2)=  
    \begin{cases}
         0,  & \mathsf{LCA}(x_1, \, x_2) \,\, \textrm{is leaf}; \\
        e^{-\lambda \cdot \mathsf{depth}(\mathsf{LCA}(x_1, x_2))}, & \text{otherwise},
\end{cases}
\label{eq:lca}
\end{equation}
where \textsf{LCA} refers to the {\em least common ancestor} of the examples $x_1$ and $x_2$. Figure \ref{fig:lca} shows examples $x_1 \equiv$ {\tt advisedBy(Tom, Mary)} and $x_2 \equiv$ {\tt AdvisedBy(Tom, John)}; they both follow the same path down the tree before diverging at a node at depth $2$. Now, consider $x_1$ and $x_3 \equiv$ {\tt AdvisedBy(Ada, Dan)}. In this case, we have that the least common ancestor is at depth $1$. Since the distance measure is inversely related to depth of the least common ancestor, we have that $x_1$ and $x_2$ are closer together than $x_1$ and $x_3$. 
Typically, more than one tree is learned (say, via functional gradient boosting), and the one-class classifier is a weighted combination of these trees. Then, the overall distance function is simply the weighted combination of the individual tree-level distances: $D(x_1, \, x_2)=\sum_i \, \beta_i \, d_i(x_1, x_2)$
where $\beta_i$ is the weight of the $i^{th}$ tree and $\sum_i \beta_i=1, \beta_i \geq 0$. 
The non-parametric function $D(\cdot, \cdot)$ is a relational distance measure learned on the data. The distance function can then be used to compute the density estimate for a new relational example $z$ as a weighted combination of the distance of $z$ from all training examples $x_j$, $E(z \not \in \texttt{class}) \, = \, \sum_j\, \alpha_j D(x_j, \,z)$, 
where $\alpha_j$ is the weight of the labeled example $x_j$ and  $\sum \alpha_j=1, \alpha_j \geq 0$. Note that expectation above is for $z \not \in \texttt{class}$, since the likelihood of class membership of $z$ is inversely proportional to its distance from the training examples describing  that \texttt{class}.

We learn a tree-based distance iteratively \cite{khot2014relational} to introduce new relational features that perform one-class classification. The left-most path in each relational tree is a conjunction of predicates, that is, a clause, which can be used as a relational feature. 
The splitting criteria is the squared error over the examples and the goal is to minimize squared error in each node as shown in equation \ref{eq:obj}. 
\begin{footnotesize}
\begin{equation}
\begin{array}{l}{\min \sum_{y \in \mathbf{x}_{\mathrm{r}}}\left[I(z)-E(z \notin \texttt{class})-\Sigma_{j : x_{j} \in \mathbf{x}_{l}} \alpha_{j} \beta_{i} d_{i}\left(x_{j}, z\right)\right]^{2}} \\ {+\sum_{y \in \mathbf{x}_{l}}\left[I(z)-E(z \notin \texttt{class})-\Sigma_{j : x_{j} \in \mathbf{x}_{r}} \alpha_{j} \beta_{i} d_{i}\left(x_{j}, z\right)\right]^{2}}\end{array}
\label{eq:obj}
\end{equation}
\end{footnotesize}
$I(z)$ is the indicator function and returns 1 if $z$ is an unlabeled example or 0 otherwise. Also, $x_l$ and $x_r$ are the examples that take the left and right branch respectively. A greedy search approach is employed  for tree learning, thereby providing a \textit{non-parametric} approach for learning these relational trees. Algorithm \ref{alg:struct} shows our structure learning method for DGMs using relOCC.

\begin{algorithm}
  \caption{\small Structure Learning using relOCC; {\bf Input}: fact base $\mathbf{F_s}$, positive ex. $\mathsf{pos}$, negative ex. $\mathsf{neg}$}
  \label{SLR}
  {\small
  \begin{algorithmic}[1]
    \Function{$\mathtt{LearnGaifmanStruct}$}{$\mathbf{F_s},\, \mathsf{pos}, \, \mathsf{neg}$}
   \For{{\bf every} $\mathbf{x_1,x_2}$ {\bf in} $\mathsf{pos}$}
    \State Calculate $d(x_1,x_2)$ according to equation \ref{eq:lca}
    \State $D(x_1, \, x_2)=\sum_i \, \beta_i \, d_i(x_1, x_2)$
    \Comment{Compute weighted distance between $x_1,x_2$}
    \EndFor
    \For{a given new unlabeled example $\mathbf{z}$}
    \State $E(z \not \in \texttt{class}) \, = \, \sum_j\, \alpha_j D(x_j, \,z)$ 
    \Comment{Calculate the density estimate}
    \EndFor
    \State Learn the tree $\mathtt{T}$ iteratively by minimizing equation \ref{eq:obj}
    \State {\bf return} $\mathtt{LeftBranch}(\mathtt{T})$
\EndFunction
\end{algorithmic}}
\label{alg:struct}
\end{algorithm}

\begin{algorithm}[h]
  \caption{\small Learning Embeddings from Discriminative Gaifman Models; {\bf Input}: target query $\mathbf{q}$, knowledge base $\mathcal{B}$, positives $\mathsf{pos}$, negatives $\mathsf{neg}$; {\bf Params}: depth $r$, size $k$ and number of Gaifman neighborhoods $w$}
  \label{SL}
  {\small
  \begin{algorithmic}[1] 
    \Function{$\mathtt{LGE}$}{$\mathbf{q}, \, \mathcal{B},\, \mathsf{pos}, \, \mathsf{neg}$}
  \State $\mathcal{G} =  \mathtt{MakeGaifmanGraph(\mathcal{B})}$  \Comment{construct Gaifman graph from facts}
  \State $F_s = \mathtt{MakeFactBase(\mathcal{B})}$ 
  \State  $\Phi = \mathtt{LearnGaifmanStruct}(F_s, \, \mathsf{pos}, \, \mathsf{neg})$ \Comment{extract relational features (section \ref{subsec: relational features})}
    \State  $G_{pos}, \, G_{neg} = \mathtt{Ground}(\Phi, \, F_s, \, \mathsf{pos})$ \Comment{ground positive and negative examples}
    \State $T_\mathbf{q}^\mathsf{pos}, \, T_\mathbf{q}^\mathsf{neg} = \mathtt{GetQueryTuples(\mathbf{q}, \, F_s)}$  \Comment{all tuples satisfying $\mathbf{q} \in F_s$ (pos), $\neg \mathbf{q} \in F_s$, (neg) }    
  \For{{\bf every} $\mathbf{t}$ {\bf in} $T_\mathbf{q}^\mathsf{pos}$} 
        \State  $\mathcal{N} = \mathtt{GenerateNeighborhoods}(\mathbf{t}, \, r, \, k, \, w)$ \Comment{generate $w$ neighborhoods of depth $r$ and size $k$}
        \For{{\bf every} $\varphi$ {\bf in} $\Phi$}
            \State $\thetab = \varphi / \mathbf{t}$ \Comment{substitute query tuple $\mathbf{t}$ in feature $\varphi$}
            \State $x^\varphi_\mathbf{t} \, = \, \mathtt{Count}(\thetab, \, \mathcal{N}, \, G_{pos})$  \Comment{count groundings satisfied in the neighborhoods}
        \EndFor
        \State $\mathbf{x}_\mathbf{t}^\mathsf{pos} = [\hdots, \, x^\varphi_\mathbf{t}, \, \hdots, \, x_{|\Phi|}]$ \Comment{embedding for tuple $\mathbf{t}$}
    \EndFor
    \For{{\bf every} $\mathbf{t}$ {\bf in} $T_\mathbf{q}^\mathsf{neg}$} 
        \State  $\mathcal{N} = \mathtt{GenerateNeighborhoods}(\mathbf{t}, \, r, \, k, \, w)$ \Comment{generate $w$ neighborhoods of depth $r$ and size $k$}
        \For{{\bf every} $\varphi$ {\bf in} $\Phi$}
            \State $\thetab = \varphi / \mathbf{t}$ \Comment{substitute query tuple $\mathbf{t}$ in feature $\varphi$}
            \State $x^\varphi_\mathbf{t} \, = \, \mathtt{Count}(\thetab, \, \mathcal{N}, \, G_{neg})$ \Comment{count groundings satisfied in the neighborhoods}
        \EndFor
        \State $\mathbf{x}_\mathbf{t}^\mathsf{neg} = [\hdots, \, x^\varphi_\mathbf{t}, \, \hdots, \, x_{|\Phi|}]$ \Comment{embedding for tuple $\mathbf{t}$}
    \EndFor
\State {\bf return} $\F= \{\mathbf{x}_\mathbf{t}^\mathsf{pos}\}, \, \{ \mathbf{x}_\mathbf{t}^\mathsf{neg}\}$ \Comment{return embeddings}
\EndFunction
\end{algorithmic}}
\label{alg:method}
\end{algorithm}

\subsection{Feature Construction}
Once extracted, relational rules are instantiated (grounded) to obtain graphs $G_{pos}$ and $G_{neg}$. While several {\em feature aggregations} exist, we employ {\em counts} since they have been successfully employed in many relational models. For every relational feature $\varphi \in \Phi$, the first and last entity are instantiated 
corresponding to the tuples satisfying the query.
For example, consider the knowledge base snippet in Fig. \ref{fig:gaif}; let the positive example be {\tt Interacts(Pravastatin, Simvastatin)}. For a relational feature, say {\tt EnzymeInhib(d0, t0)} $\wedge$ {\tt \_EnzymeInhib(t0, d1)}, and the substitution $\{\mathtt{d0}/\mathtt{Pravastatin}, \mathtt{d1}/\mathtt{Simvastatin} \}$ 
we obtain the {\bf partially-grounded relational feature} {\tt EnzymeInhib(Pravastatin, t0)} $\wedge$ {\tt \_EnzymeInhib(t0, Simvastatin)}. 
Next, all the entities that completely satisfy this partially grounded feature are obtained. The features for each query variable are then obtained as counts of the number of entities in the satisfied grounded features that are also present in the neighborhood of the query entities in the Gaifman graph $\mathcal{G}$. For example, in the Gaifman graph in Figure \ref{fig:gaif}, we check if {\tt EnzymeInhib(Pravastatin, CytochromeP4502C9)} $\wedge$ {\tt \_EnzymeInhib(CytochromeP4502C9, Simvastatin)} satisfies the given relational feature i.e. this grounding $ \in G $. If the grounding satisfies the relational feature and since {\tt CytochromeP4502C9} is present in the Gaifman neighborhood of {\tt Pravastatin} (as well as {\tt Simvastatin}), the count of the relational feature is increased by $1$. Thus, for every query variable $q$ we obtain a propositional feature $f=[f_1,....,f_{|\Phi|}]$ of length $|\Phi|$:
\begin{equation}
    f_i=
    \begin{cases}
     |\psi^{N_r(q)}(q)|,  &\text{if $q(e_1,e_2)$ partially grounds $\Phi_i$, }\\
     0, & \text{otherwise}.
\end{cases}
\end{equation}
Recall that $\psi$ refers to the relativized first-order formula, and consequently $\psi^{N_r(q)}(q)$ is the $r$-local formula for a neighborhood $N$ of depth $r$. Thus, we count the number of entities in the satisfied grounded features that are also satisfied in the neighborhood structure of the Gaifman graph.

\begin{table*}[h]
    \centering
    \caption{Evaluation domains and their properties.}
    \label{tab:doms}
    \scalebox{1.0}{
    \begin{tabular}{|l|c|c|c|c|c|c|c|}
    \hline
       \textbf{Data set}  &  \textbf{\#Entities} & \textbf{\#Relations} & \textbf{\#Pos} & \textbf{\#Neg}&  \textbf{\#RW rules} & \textbf{\#ILP rules} & \textbf{\#relOCC rules}\\
       \hline
       \hline
        DDI & $355$ & $15$ & $2832$ & $3188$ & $68$ & $36$ & $25$\\
        PPI & $797$ & $7$ & $1915$ & $1915$ & $42$ & $5$ & $15$\\
        NELL Sports & $4147$ & $6$ & $300$ & $600$ & $36$ & $15$ & $13$\\
        Financial NLP & $650$ & $7$ & $186$ & $1029$ & $222$ & $6$ & $25$\\
        ICML Co-Author & $558$ & $5$ & $155$ & $6498$ &  $7$ & $15$ & $7$\\
        \hline
    \end{tabular}}
    \vspace{-0.2in}
\end{table*}

Procedure \ref{alg:method} presents our method, LGE for extracting embeddings from DGMs. In \textbf{[Line 2--3]:} we build the initial Gaifman graph $\mathcal{G}$. 
In \textbf{[Line 4]:} we learn the relational features from one of the methods defined in \ref{subsec: relational features}, which are then grounded using both the positive and negative examples \textbf{[Line 5]}; in addition, tuples of the positive and negative examples are also obtained \textbf{[Line 6]}. 
For both positive ($T^\mathsf{pos}_\mathbf{q}$) and negative tuples ($T^\mathsf{pos}_\mathbf{q}$), the neighborhood of each entity in the tuple is obtained, and each relational feature is partially grounded with the entities $\in t$ \textbf{[Line 8, 16]}. {\tt GenerateNeighbors} \cite{niepert2016discriminative} generates entity neighborhoods for a tuple $\mathbf{t} \in T_\mathbf{q}$. 
Neighborhood generation relies on three parameters: (1) $r$, the depth of neighborhood when counting, (2) $k$, the number of neighbors to sample, and (3) $w$, the number of neighborhoods to be generated. 
For each entity in tuple ${t}$, all neighbors at a maximum distance of $r$ form the neighborhood (Fig. \ref{fig:neigh}, the outer region). This process is repeated until we obtain $w$ neighborhoods for each training example. For example, if $r=1, w=5$ and $k=10$ and we have 10 relational features ($| \Phi | = 10$), we obtain 50 propositional examples with 10 features by looking at $1$-neighbors for each entity.
The $\mathtt{Count}$ function \textbf{[Line 11, 19]} counts how many entities in the neighborhood of each query satisfy the partially-grounded relational features. Each such count becomes a propositional feature. In this manner, we can construct a propositionalized data set of $k \times w$ positive examples and $k \times w$ negative examples.

\textbf{Learning a Discriminative Model:} After learning the propositional features, any standard classifier can be used for link prediction. 
In our experiments, we employ gradient-boosting~\cite{friedman2001greedy} and logistic regression. Results using more algorithms are given in the Appendix. The classification algorithm itself is not a key contribution of our work and as we demonstrate empirically next, a standard classifier suffices for learning an effective discriminative model. 

\begin{table*}[t]
    \centering
    \caption{Results for the relational domains. Note that the first three data sets are relatively balanced and the last two are unbalanced. Thus, we do not report Accuracy and AUC-ROC and instead report AUC-PR for the last 2 data sets.}
    \label{tab:res_bal}
    \begin{tabular}{c|c|cc|cc|cc|cc|cc}
        \hline
        \small
         Data set &  Methods & \multicolumn{2}{c|}{Accuracy} & \multicolumn{2}{c|}{Recall} & \multicolumn{2}{c|}{F1} & \multicolumn{2}{c|}{AUC-ROC} & \multicolumn{2}{c}{AUC-PR}\\
         & & LR & GB & LR & GB & LR & GB & LR & GB & LR & GB\\
        \hline
        \multirow{3}{*}{DDI} & RW & 0.657 & 0.669 & 0.469 & 0.530 & 0.564 & 0.602 & 0.647 & 0.662\\
        & ILP  & 0.696 & 0.774 & 0.467 & 0.674 & 0.592 & 0.729 & 0.684 & 0.767\\
        & relOCC & 0.860 & \textbf{0.897} & 0.939 & \textbf{0.991} & 0.864 & \textbf{0.901} & 0.864 & \textbf{0.902}\\
        & MLN-Boost & \multicolumn{2}{c|}{0.711} & \multicolumn{2}{c|}{0.504} & \multicolumn{2}{c|}{0.618} & \multicolumn{2}{c|}{0.798}\\
        \hline
        \multirow{3}{*}{PPI} & RW & 0.700 & \textbf{0.785} & 0.586 & 0.707 & 0.661 & 0.767 & 0.699 & \textbf{0.785}\\
        & ILP  & 0.613 & 0.661 & 0.397 & 0.553 & 0.506 & 0.620 & 0.613 & 0.661\\
        & relOCC & 0.727 & 0.733 & 0.996 & \textbf{0.999} & 0.785 & \textbf{0.789} & 0.727 & 0.733\\
        & MLN-Boost & \multicolumn{2}{c|}{0.649} & \multicolumn{2}{c|}{0.453} & \multicolumn{2}{c|}{0.571} & \multicolumn{2}{c|}{0.743}\\
        \hline
        \multirow{3}{*}{NELL Sports} & RW & 0.783 & 0.822 & 0.414 & 0.569 & 0.569 & 0.689 & 0.696 & 0.762\\
        & ILP  & 0.782 & 0.824 & 0.431 & 0.590 & 0.578 &  0.699 & 0.699 & 0.769\\
        & relOCC & 0.793 & \textbf{0.833} & 0.431 & 0.6 & 0.59 &  0.714 & 0.708 & {0.778}\\
        & MLN-Boost & \multicolumn{2}{c|}{0.822} & \multicolumn{2}{c|}{0.533} & \multicolumn{2}{c|}{0.667} & \multicolumn{2}{c|}{\textbf{0.894}}\\
        \hline
        \multirow{3}{*}{Financial NLP}& RW & & & 0.0 & 0.0 & 0.0 & 0.0 & & & 0.168 & 0.168 \\
        & ILP  & & & 0.068 & 0.633 & 0.112 & 0.727 & & & 0.200 & 0.6023\\
        & relOCC & & & 0.788 & \textbf{0.800} & 0.882 & \textbf{0.889} & & & 0.826 & \textbf{0.833}\\
        & MLN-Boost & & & \multicolumn{2}{c|}{0.764} & \multicolumn{2}{c|}{0.757} & & & \multicolumn{2}{c}{0.807}\\
        \hline
        \multirow{3}{*}{ICML CoAuthor} & RW & & & 0.0 & 0.0 & 0.0 & 0.0 & & & 0.023 & 0.023\\
        & ILP  & & & 0.272 & 0.339 & 0.427 & 0.506 & & & 0.289 & 0.356\\
        & relOCC & & & 0.346 & \textbf{0.386} &  0.517 &  \textbf{0.557} & & & 0.370 & \textbf{0.40}\\
        & MLN-Boost & & &\multicolumn{2}{c|}{0.326} & \multicolumn{2}{c|}{0.214} & & & \multicolumn{2}{c}{0.210}\\
             \hline
    \end{tabular}
\end{table*}

\begin{figure*}
    \centering
    \includegraphics[width=0.9\textwidth]{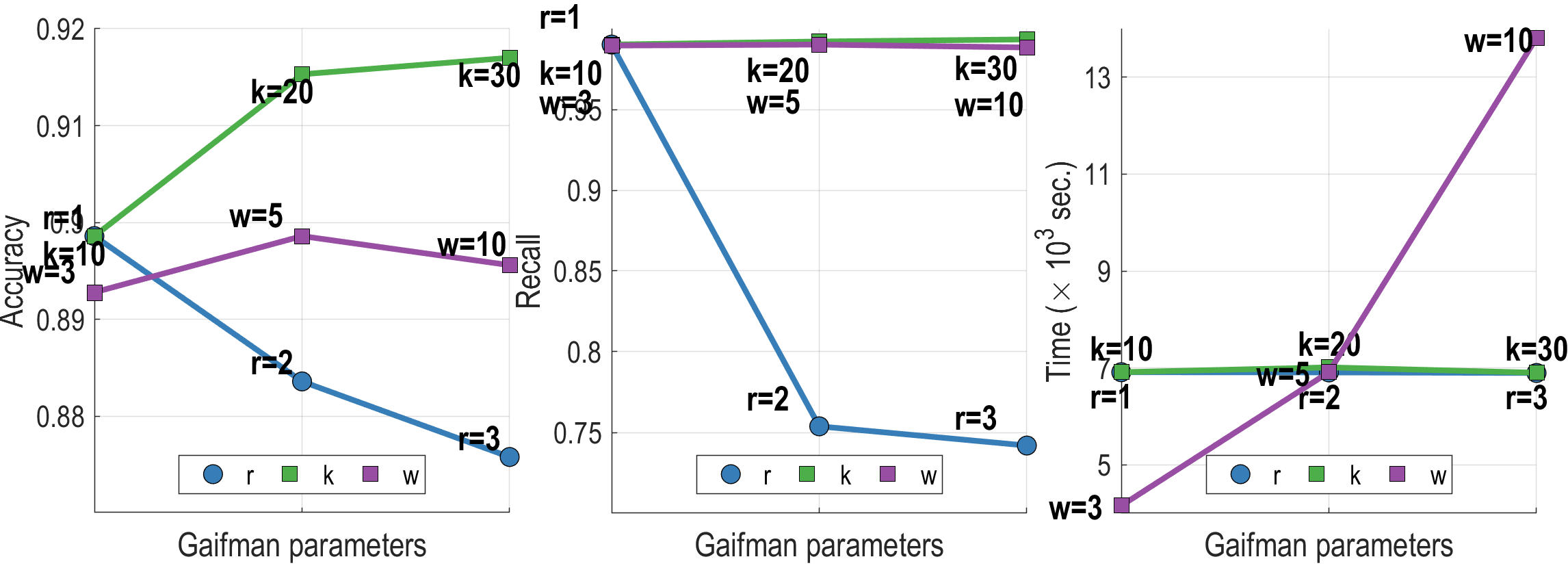}
    \vspace{-0.1in}
    \caption{\small (left) Accuracy, (middle) recall and (right) running time for various values of $r$, $k$ and $w$. For varying $r$: $w$=5 and $k$=10, for varying $w$: $r$=1 and $k$=10 and for varying $k$: $w$=5 and $r$=1.}
    \label{fig: DDI_parameters}
    \vspace{-0.2in}
\end{figure*}
\begin{figure*}[t]
\subfigure[\small Comparison of our method with Tuffy on balanced data sets.]{
        \centering
        \hspace{-0.1in}
        \includegraphics[width=0.6\columnwidth]{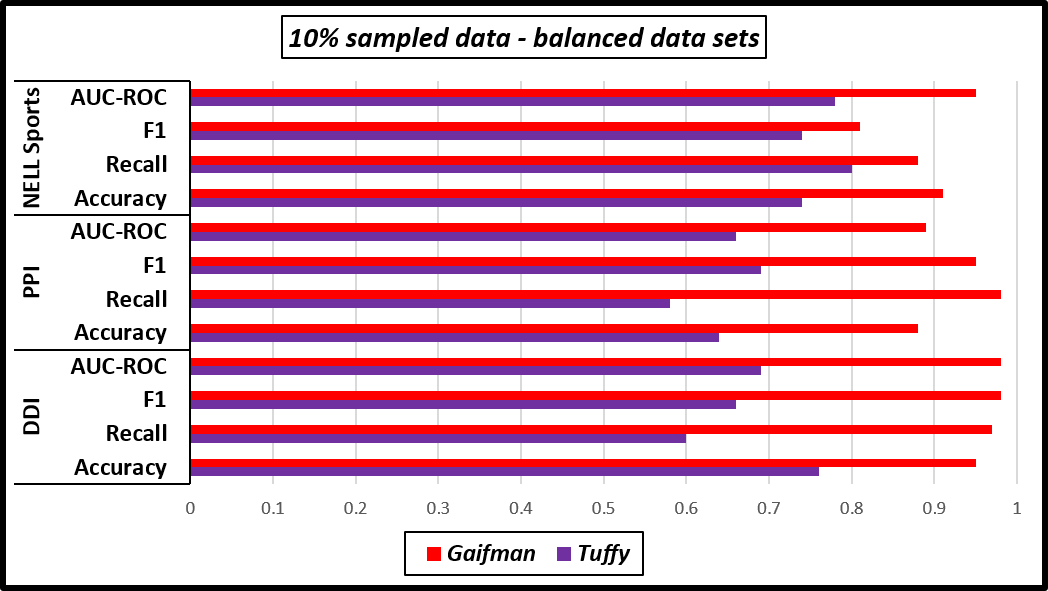}
        \label{fig:bal}
}
\hspace{0.1in}
\subfigure[\small Comparison of our method with Tuffy on unbalanced data sets.]{
        \centering
        \includegraphics[width=0.6\columnwidth]{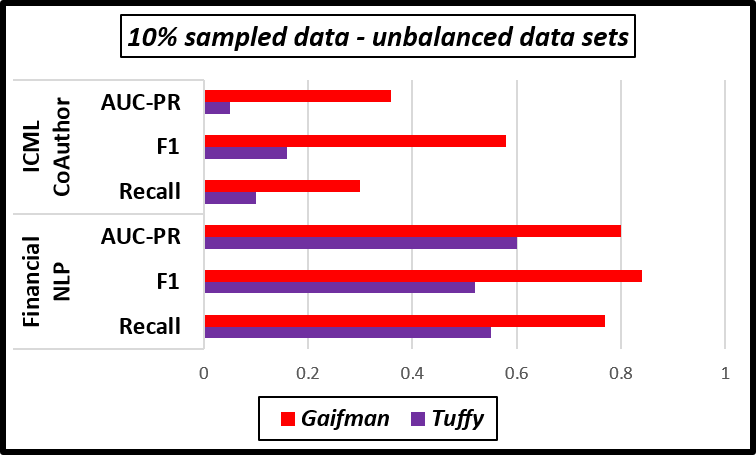}
        \label{fig:unbal}
} 
\hspace{0.1in}
\subfigure[\small Comparison of learning + inference time taken by our method with Tuffy.]{
        \centering
        \includegraphics[width=0.6\columnwidth]{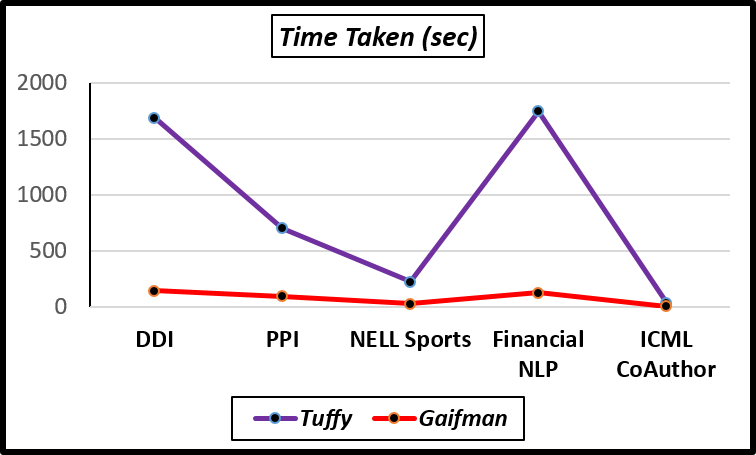}
        \label{fig:time}
} 
\caption{\small Comparison of our method with Tuffy with 10\% sampled data sets.}
\label{fig:tuff}
\vspace{-0.25in}
\end{figure*}

\section{Experiments}
We consider 5 {\em real-world novel relational data sets of varying domains} (Table \ref{tab:doms}) in our empirical evaluation 
of feature generation using random walks, ILP and relOCC. We aim to answer the following questions: \textbf{Q1:} How do different feature selection strategies compare across diverse domains from different applications? \textbf{Q2:} Does choice of the discriminative algorithm impact the performance? \textbf{Q3:} How do different feature selection strategies impact performance in domains with high class imbalance? \textbf{Q4}: What are effects of Gaifman locality parameters $r$, $w$ and $k$?  \textbf{Q5}: How does our method compare with state-of the art probabilistic ILP systems?

\textbf{Data sets:} 
\textbf{Drug-Drug interactions (DDI):} This data set~\cite{dhami2018drug} consists of $78$ drugs obtained from DrugBank and the target is the \textit{Interactions} relation between the drug entities. \textbf{Protein-Protein interactions (PPI):} This data set~\cite{kok2009alchemy} is obtained from Alchemy. The target is \textit{interaction} relation between two protein entities. \textbf{NELL Sports:} was generated by the Never Ending Language Learner (NELL) \cite{mitchell2018never} consisting of information about players and teams and the task is to  predict whether a team plays a particular sport i.e \textit{teamplayssport}. \textbf{Financial NLP:} is obtained by extracting information from \emph{SEC Form S-1} documents, which were scraped and converted into relational format and the target 
the relation \textit{sentenceContainsTarget} between sentence and word entities. \textbf{ICML Co-Author:} is obtained by mining publication data from ICML 2018; the target is the \textit{CoAuthor} relation between persons.

\textbf{Results:}
Table \ref{tab:doms} also shows the number of relational rules learned by different techniques. Table \ref{tab:res_bal} present the results for all the relational domains, after 5-fold cross validation, with logistic regression (LR) and gradient boosting (GB). All experiments were run on a \emph{64-bit Intel(R) Xeon(R) CPU E5-2630 v3} server with parameter values $r$=1, $k$=10 and $w$=5.

To answer \textbf{Q1}, we note that relOCC outperform ILP and relational RWs across a majority of the domains. This is expected since relOCC considers the density of the positive and negative examples separately, allowing the features it generates to discriminate better. In answer to \textbf{Q2}, results in Table \ref{tab:res_bal} show that choice of classifier does not make much difference in the final performance after learning relational rules, though the performance of GB is almost always higher than LR. The ICML CoAuthor (neg-to-pos ratio of 42:1) and Financial NLP (neg-to-pos ratio of 6:1) are highly imbalanced; consequently, we report AUC-PR. In both domains, AUC-PR for relOCC outperforms the other rule-construction methods by a large margin. Random walk rules, in particular, cause all the examples to be classified as negative, and results in recall and F1-score of $0$ in both the domains. Thus, to answer \textbf{Q3}: different feature selection strategies do affect the performance in highly-imbalanced domains.
To answer \textbf{Q4}, figure \ref{fig: DDI_parameters} show the effects of varying $r$ (depth of neighborhoods), $k$ (number of neighbors) and $w$ (number of neighborhoods) on the DDI data set. Generally, $k$ does not affect performance significantly, but increasing $r$ causes recall report to drop sharply. This is because, with $r=1$, entities in the query neighborhood are more tightly coupled with entities in the query variables. Another important takeaway is that the rules learned using relOCC exhibit {\em high clinically-relevant recall ($\approx$ 1) on medical data sets}: DDI and PPI. This has considerable implications for bioinformatics domains as recall is the most important metric; this is because a false negative (such as a misdiagnosis) results in much more serious consequences \cite{dhami2018drug} than a false positive. Finally, from figure \ref{fig: DDI_parameters} (right), we note that 
varying $r$ and $k$ does not change the run time. 
However, increasing $w$ increases the run time since the size of the neighborhood graph to be searched increases. To answer \textbf{Q5}, we also compare our method to two probabilitic ILP systems, MLN-Boost \cite{khot2011learning} and Tuffy \cite{niu2011tuffy}. Table \ref{tab:res_bal} shows that our method outperforms MLN-Boost by a significant margin. We also note that Tuffy \footnote{We also tried systems such as Alchemy, Problog, ProbCog} could not handle the amount of data that we have used in our learning framework and thus could not learn the structure. Instead, used the ILP rules that we learned, and tried learning the weights but Tuffy could not complete training after a few hours. Thus, we sampled 10\% data from all the data sets and the results for the same are presented in figure \ref{fig:tuff}. In case of limited number of samples as well, our method is significantly better than the PILP system.

\section{Conclusion}
We considered the problem of full model learning of discriminative Gaifman models.  Our algorithm first constructs a set of rules, identifies the appropriate instantiations and finally counts the number of groundings per rule. These become the raw features based on which one could train a discriminative classifier. Our work provides a method of constructing relational embeddings in an effective manner that can be combined with a scalable local model. Given the {\em increasing importance of local neighborhoods in graph data, automatic learning of these neighborhoods is an important direction and contribution.}
One possible future direction could be employing more graph based embedding methods that can integrate with Gaifman's locality principle. Evaluating on more databases and knowledge graphs is an interesting direction.

\section*{Acknowledgments}
\label{sec:Acknowledgments}
SN and GK gratefully acknowledge AFOSR award FA9550-18-1-0462. DD and SY gratefully acknowledge DARPA Minerva award FA9550-19-1-0391. Any opinions, findings, and conclusion or recommendations expressed in this material are those of the authors and do not necessarily reflect the view of the AFOSR, DARPA or the US government.


\bibliographystyle{aaai}
\bibliography{Gaifman}

\end{document}